\documentclass[preprint,12pt]{elsarticle}

\usepackage[colorlinks=false,citecolor=black,linkcolor=magenta]{hyperref}
\usepackage{geometry}
\usepackage{booktabs}
\usepackage{natbib}
\usepackage{algpseudocode}
\usepackage{algorithm}
\usepackage{array}
\usepackage{rotating}
\usepackage{graphicx}
\usepackage{multirow}
\usepackage{setspace}
\usepackage{subcaption}
\usepackage{amsmath}
\usepackage{threeparttable}
\usepackage{color}
\usepackage{amssymb}
\usepackage{breqn}
\usepackage{lscape}
\usepackage[normalem]{ulem}
\usepackage{soul}

\usepackage{enumerate}
\usepackage[shortlabels]{enumitem}

\newcolumntype{P}[1]{>{\centering\arraybackslash}p{#1}}
\newcolumntype{Q}[1]{>{\raggedleft\arraybackslash}p{#1}}
\newcolumntype{C}[1]{>{\centering\arraybackslash}p{#1}}

\journal{Reliability Engineering and System Safety}
\usepackage{lineno}

\usepackage{xcolor}
\renewcommand\fbox{\fcolorbox{black!20}{white}}
\makeatletter
\def\ps@pprintTitle{%
	\let\@oddhead\@empty
	\let\@evenhead\@empty
	\def\@oddfoot{}%
	\let\@evenfoot\@oddfoot}

\makeatother

\begin{document}
	
	\begin{frontmatter}
		\title{Application of Clustering Algorithms for Dimensionality Reduction in Infrastructure Resilience Prediction Models}
		
		
		\author{Srijith Balakrishnan\corref{correspondingauthor}}
		\address{Postdoctoral Researcher, Future Resilient Systems, Singapore-ETH Centre at CREATE, 1~Create Way, Singapore 138602}
		\author{Beatrice Cassottana\corref{cor3}}
		\address{Postdoctoral Researcher,  Future Resilient Systems, Singapore-ETH Centre at CREATE, 1~Create Way, Singapore 138602}
		\author{Arun Verma\corref{cor3}}
		\address{Postdoctoral Fellow,  Department of Computer Science, National University of Singapore, COM1, 13, Computing Dr, Singapore 117417}
		\cortext[correspondingauthor]{Corresponding author. Email: \href{mailto:srijith.balakrishnan@sec.ethz.ch}{srijith.balakrishnan@sec.ethz.ch}, Tel: +65 98063679}
		
		\thispagestyle{empty}
		\begin{abstract}
			Recent studies increasingly adopt simulation-based machine learning (ML) models to analyze critical infrastructure system resilience. For realistic applications, these ML models consider the component-level characteristics that influence the network response during emergencies. However, such an approach could result in a large number of features and cause ML models to suffer from the `curse of dimensionality'. We present a clustering-based method that simultaneously minimizes the problem of high-dimensionality and improves the prediction accuracy of ML models developed for resilience analysis in large-scale interdependent infrastructure networks. The methodology has three parts: (a) generation of simulation dataset, (b) network component clustering, and (c) dimensionality reduction and development of prediction models. First, an interdependent infrastructure simulation model simulates the network-wide consequences of various disruptive events. The component-level features are extracted from the simulated data. Next, clustering algorithms are used to derive the cluster-level features by grouping component-level features based on their topological and functional characteristics. Finally, ML algorithms are used to develop models that predict the network-wide impacts of disruptive events using the cluster-level features. The applicability of the method is demonstrated using an interdependent power-water-transport testbed. The proposed method can be used to develop decision-support tools for post-disaster recovery of infrastructure networks.
		\end{abstract}
		
		
	\end{frontmatter}
	\par
	
	
	
	
	
	\newpage
	
	\section{Introduction}
	The intensifying natural disasters and emergence of new threats, such as cyber-attacks and pandemics, have led to a paradigm shift towards infrastructure system resilience. The physical and functional risks posed by such extreme events to our infrastructure systems are compounded by climate change, the drastic modifications to the built environment, and the growing interdependence among urban systems. Therefore, the infrastructure systems are no longer designed for operational efficiency alone; equal emphasis is given to their ability to withstand disasters and minimize the resultant societal and economic impacts from unanticipated service disruptions \citep{Gay2013}. Disaster mitigation and resilience enhancement alternatives have become crucial aspects considered in the design and management of existing infrastructure systems.
	
	Several approaches exist to analyze the resilience capabilities of interdependent infrastructure systems and evaluate the resilience project alternatives. These approaches can be broadly classified into two, namely empirical- and computational approaches \citep{Mitsova2021}. Empirical approaches rely on datasets, records, and reports based on historical events to identify patterns and severity in physical and functional disruptions to infrastructure systems \citep{McDaniels2007, Luiijf2009}. Data collected during historical breakdown events are used to characterize interrelationships among different infrastructure systems. On the other hand, computational approaches attempt to replicate the physical-, cyber-, geographic-, and logical dependencies among infrastructure components using mathematical and logical functions. Several models, such as, network theory-based models \citep{Holden2013,Praks2017}, system dynamics models \citep{Pasqualini2005,Powell2008}, agent-based models \citep{Tesfatsion2003,Thompson2019} and input-output models \citep{Haimes2005,Oliva2010}, have been extensively used for modeling interdependent infrastructure systems and analyzing their resilience. More advanced computational models adopt co-simulation of multiple domain-specific infrastructure simulators instead of a homogeneous method to replicate the collective behavior of interdependent infrastructure systems \citep{Wang2022}.
	
	Recent computational approaches have been increasingly adopting simulation-based machine learning (ML) models to predict the network-wide impacts and use these predictions for the analysis of interdependent infrastructure resilience \citep{Golkhandan2021}. Most of the ML models for infrastructure resilience analysis focus on the optimal allocation of resources for improving the absorptive and recovery capabilities of infrastructure systems. \cite{Alemzadeh2020} developed Artificial Neural Network (ANN) models to approximate post-disaster repair sequences for optimal recovery of interdependent infrastructure systems by training simulated data. \cite{Sun2020} applied a Deep Q-learning (DQN) algorithm on an interdependent power-water-transport model to predict the optimal repair crew allocation to flood-affected bridges that would minimize the cumulative impact on the network. \cite{Dehghani2021} applied Deep Reinforcement Learning (DRL) to identify the optimal long-term preventive maintenance strategy that maximizes the resilience of power systems.
	
	Although ML models are powerful tools to accurately predict infrastructure system resilience (i.e., network-wide impacts of different disruptive events), they require large training datasets to ensure adequate prediction accuracy. Most of the aforementioned ML models learn the vulnerability and resilience attributes at a component level, leading to a large number of features in the model, also known as the `\textit{curse of dimensionality}' \citep{Turati2016}. When the dimensionality of a problem increases, generalizing the trends becomes harder because the training dataset of fixed size can cover only a small fraction of the possible input combinations \citep{Domingos2012}. While using a larger simulation dataset could be a potential solution to the problem of high-dimensionality, several advanced infrastructure simulation models are computationally intensive and may require considerable time for simulating even a small number of disaster scenarios \citep{Liu2019, Zou2021}. The consequences of high dimensionality could be even more severe when the system response of large-scale interdependent infrastructure networks is to be learned by the ML algorithms.
	
	To mitigate the negative effects of high dimensionality, our study exploits clustering methods to identify similar infrastructure components in terms of their topological and functional properties and later incorporates that information to enhance the accuracy of resilience prediction models. Several studies in the literature have demonstrated that topological and functional attributes of components influence the infrastructure network vulnerability and resilience characteristics \citep{Cadini2009,Nicholson2016,Balakrishnan2020}. 
	
	The overall objective of this paper is to propose a novel network clustering-based approach for achieving dimensionality reduction in simulation-based machine learning models for infrastructure network analysis. The specific objectives are as follows:
	\begin{enumerate}
		\item Demonstrate how the relationship between network topology and infrastructure resilience, as established in the literature, can be leveraged to incorporate network structure characteristics in ML prediction models.
		\item Present a methodological framework to apply network clustering to reduce dimensionality in infrastructure resilience prediction models.
		\item Propose unsupervised and supervised methods to find the optimal number of clusters in each infrastructure system in an interdependent network.
	\end{enumerate}
	
	The rest of the paper is organized as follows: Section~\ref{sec:method} presents the methodological framework adopted in this paper; Section~\ref{sec:case_study} demonstrates the application of the methodology on a synthetic interdependent power-water-transport network; and Section~\ref{sec:conclusions} summarizes the findings and discusses the scope for further research.
	
	\section{Methodology}
	\label{sec:method}
\begin{sidewaysfigure}[htbp]
		\centering
		\includegraphics[width = 0.85\textwidth]{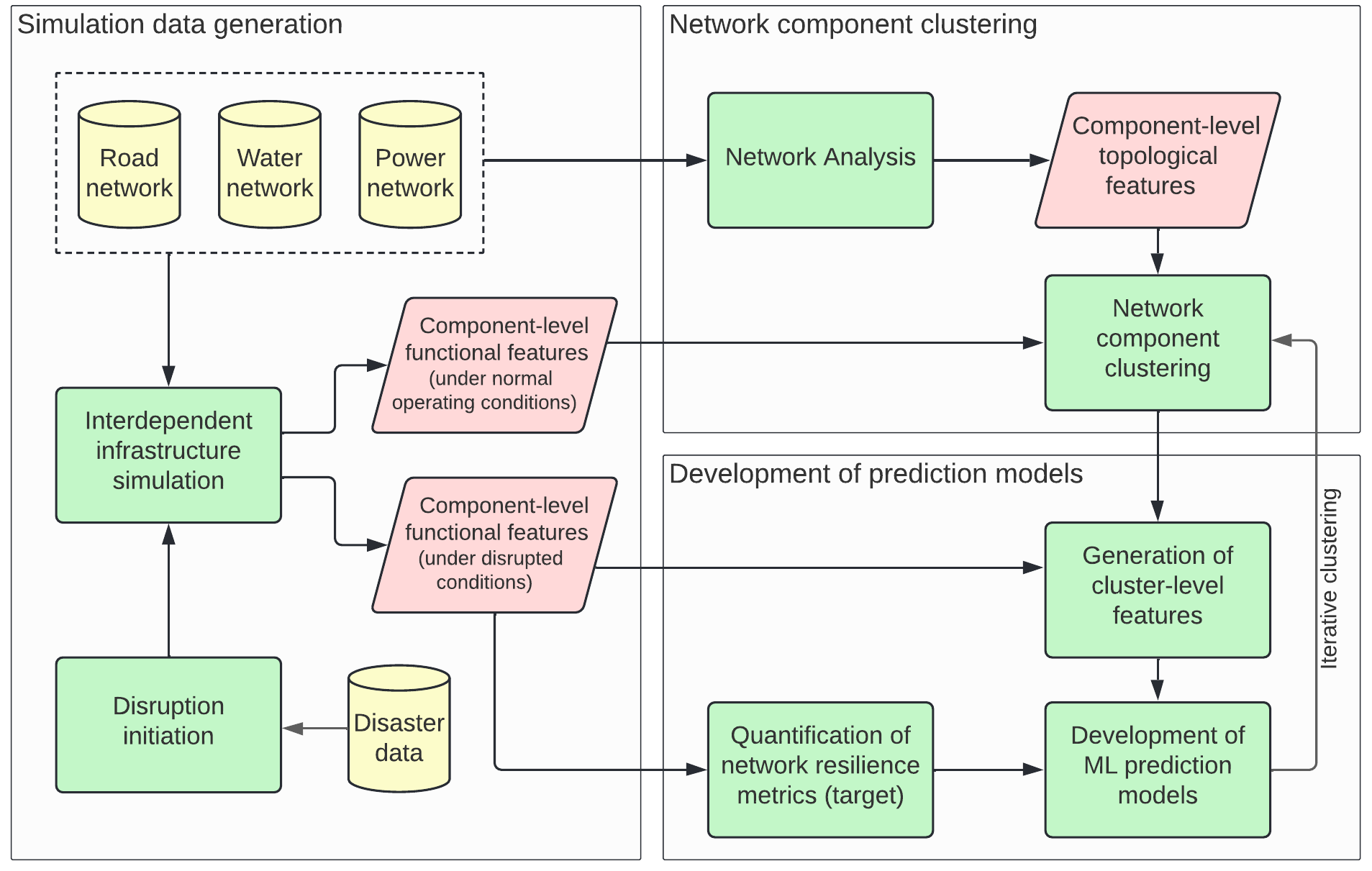}
		\caption{Framework for developing ML-based infrastructure resilience models.}
		\label{fig:methodology}
	\end{sidewaysfigure}

	The methodological framework adopted in the study is outlined in Figure~\ref{fig:methodology}. 
	The methodology is implemented in three steps, namely, (a) generation of simulation dataset, (b) network component clustering, and (c) dimensionality reduction and development of prediction models. In the first step, we employ an interdependent infrastructure model to simulate the infrastructure disruption data required for developing the prediction model. The second step is the application of appropriate clustering algorithms to categorize infrastructure components based on their topological and functional characteristics. Finally, we develop ML models for predicting the network-wide impacts using disaster and recovery-related features. To summarize, we propose algorithms that combine network component clustering with existing ML prediction algorithms to reduce dimensionality and improve model prediction simultaneously. In the rest of the section, we discuss the above steps in detail.

	\subsection{Generation of simulation dataset}
	This step derives the relevant target and predictor features using an interdependent infrastructure simulation model. The step further constitutes two subtasks: interdependent infrastructure simulation and feature extraction.
	
	\subsubsection{Interdependent infrastructure simulation}
	In this study, we employed \textit{InfraRisk}, an open-source Python-based integrated simulation platform, for simulating infrastructure disruptions and subsequent recovery in interdependent power-water-transport networks \citep{Balakrishnan2021}. \textit{InfraRisk} integrates existing domain-specific infrastructure simulators (\textit{pandapower} for power systems, \textit{wntr} for water distribution systems, and a static traffic assignment package for transport network) via an object-oriented interface to perform the interdependent simulations. All the three infrastructure simulators assign flows (power, water, or traffic) in the respective networks by minimizing the total loss subject to network-specific constraints and dependencies with other infrastructure systems.
	
	Consider an interdependent infrastructure network $\mathbb{K}$ with constituent infrastructure systems $K:K \in \mathbb{K}$. An infrastructure system $K$ can be represented as a graph $G(V_{K}, E_{K})$, where $V_{K}$ is the set of nodes and $E_{K}$ the set of links connecting the nodes. In addition, $\mathbb{K}$ also includes the dependencies between infrastructure systems, which are represented as links. We denote the set of consumers who are dependent on $\mathbb{K}$ by $N$, and the resource supply from $K$ to each consumer $n\in N$ at simulation time $t \in T$  under normal operating conditions is represented by $S_{n}^{K}(t)\in \mathbf{S}^{\mathbb{K}\times T}$.
	
	To implement the simulation, we need to define the disaster scenario and the recovery strategy for infrastructure system $K$. We can define $h:h\in H$ as the disaster scenario which results in the failure of infrastructure nodes $V_{K}^{h}\subseteq V_{K}$ and links $E_{K}^{h}\subseteq E_{K}$.
	On the other hand, the recovery strategies $P_{K}^{h}$ defines the schedule of repair actions (repair sequence and start time). Normally, each system has its own set of repair crews and restoration strategies to expedite recovery actions.
	
	The consequences of initial infrastructure disruptions due to the hazard, the network-wide functional disruptions due to infrastructure interdependencies, and the subsequent recovery efforts are reflected in the resource supplied to the consumers during the disrupted conditions $\left (s_{n}^{K}(t)\in \mathbf{s}^{\mathbb{K}\times T}\right )$ and is captured by \textit{InfraRisk}. To summarize, the interdependent infrastructure model can be represented using Equation~\ref{eq:black_box}.
	\begin{equation}
	\mathbf{s}^{\mathbb{K}\times T} = \mathbf{\Gamma}\left(\mathbb{K}, h,\mathbb{P}^h\right),
	\label{eq:black_box}
	\end{equation}
	where $\mathbf{\Gamma}(\cdot)$ is the simulation model and $\mathbb{P}^h = \{P_K^{h}: K\in \mathbb{K}\}$.

	\subsubsection{Feature extraction}
	The simulation model provides three types of data required to develop the prediction models.
	\begin{enumerate}[i.]
		\item Timeline of consumer-level and system-level functional performance under normal and disrupted network conditions.
		\item Topological characteristics of the infrastructure systems.
		\item System disruption and recovery characteristics, such as the list of initially failed components and the repair sequencing strategy used.
	\end{enumerate}
	
	Since the objective of the ML model is to predict the network-level resilience, we converted the consumer-level resource supply values generated by the simulation model to network-level resilience metrics. In this study, we used Prioritized Consumer Serviceability (PCS) as the measure of performance (MOP) for tracking the performance of the networks \citep{Balakrishnan2021}, defined as
	
	\begin{equation}
	\text{PCS}_{K}(t) = \left( \frac{\sum_{\forall i: S_{n}^{K}(t) > 0} s_{n}^{K}(t)}{\sum_{\forall i: S_{n}^{K}(t)> 0}S_{n}^{K}(t)}  \right), \text{where~} 0 \leq s_{n}^{K}(t) \leq S_{n}^{K}(t).
	\label{eq:pcs}
	\end{equation}
	
	The resource supply under normal operating conditions $S_{n}^{K}(t)\in \mathbf{S}^{\mathbb{K}\times T}$ is computed by performing the interdependent infrastructure simulation without failing any infrastructure components ($\mathbf{S}^{\mathbb{K}\times T} = \mathbf{\Gamma}(\mathbb{K})$).
	The resilience of infrastructure systems is quantified using the concept of equivalent outage hours (EOH). The EOH corresponding to each infrastructure network for a given disaster scenario $\gamma_{K}$ is given by Equation~\ref{eq:system_eoh}.
	
	\begin{equation}
	\gamma_{K} = \frac{1}{3600}\int_{t_{0}}^{t_{max}} \left [ 1 - PCS_{K}(t)\right ]dt,
	\label{eq:system_eoh}
	\end{equation}
	where $t_{0}$ is the time of occurrence of the disaster event in the simulation and $t_{max}$ is the maximum simulation time (both in seconds). Mathematically, $[1-PCS_{K}(t)]$ is the mean unmet demand (slack demand) in $K$ at time $t$ and $\gamma_{K}$ is the area of the resilience triangle \citep{Bruneau2003} formed by the PCS curve. The unit of $\gamma_{K}$ is system performance-hours.
	
	The target variable for the ML model is the resilience metric of the interdependent infrastructure network, computed as the weighted equivalent outage hours ($\overline{\gamma}$) of individual infrastructure systems (Equation~\ref{eq:wEOH}).
	
	\begin{equation}
	\overline{\gamma} = \sum_{K\in \mathbb{K}}w_{K}\gamma_{K}
	\label{eq:wEOH}
	\end{equation}
	where $w_{K}$ is the weight assigned to system $K$.
	In addition to the model inputs ($V_{K}^{h}$, $E_{K}^{h}$, and $P_K^{h}$) and outputs ($\mathbf{s}^{\mathbb{K}\times T}$, $\mathbf{S}^{\mathbb{K}\times T}$, and $\overline{\gamma}$), we also extracted the topological features of infrastructure components, such as, centrality values, from the infrastructure network to aid network clustering. 
	
	\subsection{Infrastructure component clustering}
	Clustering is an unsupervised learning method that attempts to identify the most natural way of partitioning a dataset based on similarity and dissimilarity among observations \citep{Xu2015}. Clustering could reveal the underlying structure of the dataset, which could be used to build supervised learning models with simple features and a better prediction accuracy \citep{Trivedi2015}. In this study, clustering is proposed to identify infrastructure components that are similar in their vulnerability and resilience characteristics. For this purpose, we investigate the use of topological and functional features of infrastructure components that are identified as indicators of infrastructure vulnerability, criticality, and resilience in the literature.
	
\begin{figure}[htbp]
		\centering
		\begin{subfigure}[h]{0.475\textwidth}
			\centering
			\includegraphics[width=\textwidth, trim = {0.05cm 0.05cm 0cm 0cm}, clip]{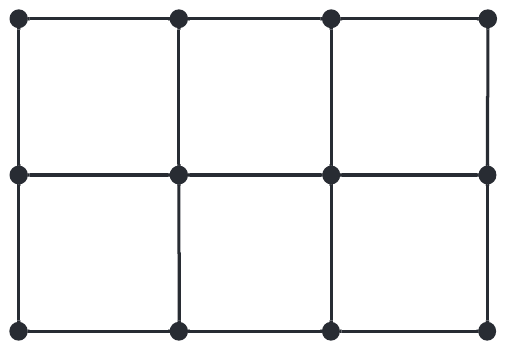}
			\caption{Original network with no clustering}
		\end{subfigure}%
		\begin{subfigure}[h]{0.475\textwidth}
			\centering
			\includegraphics[width = \textwidth, trim = 0.05cm 0.05cm 0cm 0cm, clip]{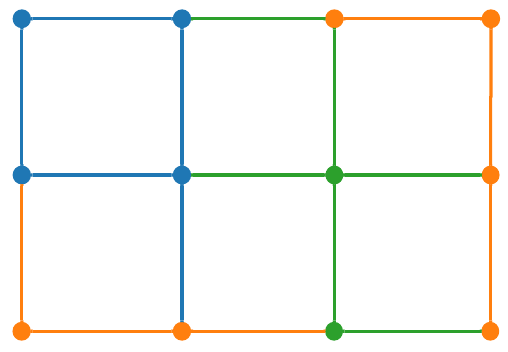}
			\caption{Clustered network}
		\end{subfigure}
		\caption{Sample illustration of clustering infrastructure network components}
		\label{fig:cluster_concept}
	\end{figure}

	Figure~\ref{fig:cluster_concept} illustrates the clustering concept introduced in the study for partitioning infrastructure components. Any infrastructure network can be treated as a set of nodes and links, where nodes represent producers, consumers, or intermediate transfer points, whereas the links denote the connections and interdependencies among the infrastructure nodes \citep{Svendsen2007, Svendsen2008}. Therefore, each component in the system can be assigned a group (or cluster) based on its topological and functional properties (color-coded in Figure~\ref{fig:cluster_concept}).
	
	A number of algorithms are available for performing clustering of datasets, such as, K-Means, K-Medoids, agglomerative propogation, and DBSCAN, depending upon the types of the datasets. For systematic reviews of clustering algorithms and their applications, see \cite{Celebi2015} and \cite{Rodriguez2019}.
	
	Consider the infrastructure system $K\in\mathbb{K}$ which can be represented as a graph $G(V_{K},E_{K})$. Without clustering, we can incorporate a characteristic associated with the infrastructure components (for example, post-disaster functional status) in two different ways as follows:
	\begin{enumerate}
		\item One feature for each component so that there will be $|V_{K}| + |E_{K}|$ additional features corresponding to each infrastructure system $K$ in the ML prediction model.
		\item A single feature which aggregates the component-level values using mathematical operations, such as summation, count, average, maximum, and minimum.
	\end{enumerate}
	
	The above two approaches have their own advantages and limitations. While the first approach may effectively learn from the spatial and structural aspects of the characteristic to be considered and ensure a high level of prediction accuracy, it leads to the issue of high-dimensionality. For a fully connected graph with $m$ nodes, $m + m(m-1)/2$ features for every component-level characteristic need to be constructed in this approach if both node- and link-level information are to be used in the ML model. In the second approach, the number of features could be considerably lower than in the first approach; however, aggregation would lead to the loss of useful information related to components, leading to a low prediction accuracy. The clustering approach aims at finding a middle ground between the above two extreme approaches. By employing appropriate clustering algorithms, we can identify $\ell_{K}^{V}:1\leq \ell_{K}^{V}\leq |V_{K}|$ node clusters and $\ell_{K}^{E}:1\leq \ell_{K}^{E}\leq |E_{K}|$ link clusters with similar functional and topological properties in the infrastructure network $K$.
	
	Several studies have demonstrated that centrality measures \citep{Cadini2009,Dunn2013,Balakrishnan2020}, such as degree centrality, betweenness centrality, and eigenvector centrality, could capture the vulnerability and resilience of infrastructure network components. In addition, many studies have shown that specific functional properties, such as flow rates under normal operating conditions, could serve as indicators of the vulnerability and importance of infrastructure components in a system \citep{Nicholson2016}. To incorporate node-specific and link-specific characteristics, clustering of nodes and links are done separately. Table~\ref{tab:topo_funcs} enlists the potential topological and functional features that are considered in the current study for network component clustering.
	
	\begin{sidewaystable}
		\begin{threeparttable}[htbp]\small
			\caption{Topological and functional features used for infrastructure component clustering}
			\begin{tabular}{p{4cm}p{5.5cm}p{8cm}}
				\toprule
				Feature   & Node component $(i)$ & Link component $(i,j)$ \\
				\midrule
				Degree centrality\tnote{1}        &   
				$\begin{aligned}
				& C_{d}(i)=\sum_{v \in V}a_{i,v}
				\end{aligned}$ & 
				$\begin{aligned}
				& C_{d}(i)=\sum_{v \in V}a_{i,v}; &
				& C_{d}(j)=\sum_{v \in V}a_{j,v}  
				\end{aligned}$\\
				Betweenness centrality\tnote{2}     & 
				$\begin{aligned}
				C_{b}(i) = \frac{\sigma_{st}(i)}{\sigma_{st}}
				\end{aligned}$ & 
				$\begin{aligned}
				& C_{b}(ij) = \frac{\sigma_{st}(i,j)}{\sigma_{st}}
				\end{aligned}$\\
				Eigenvector centrality\tnote{3}     & 
				$\begin{aligned}
				& C_{e}(i) = \frac{1}{\lambda}\sum_{v \in V}a_{i,v}C_{e}(v)
				\end{aligned}$ & 
				$\begin{aligned}
				& C_{e}(i) = \frac{1}{\lambda}\sum_{v \in V}a_{i,v}C_{e}(v); 
				& C_{e}(j) = \frac{1}{\lambda}\sum_{v \in V}a_{j,v}C_{e}(v)
				\end{aligned}$\\
				Closeness centrality\tnote{4}       & 
				$\begin{aligned}
				& C_{c}(i) = \frac{|V|-1}{\sum_{v \in V}d(i,v)}
				\end{aligned}$ & 
				$\begin{aligned}
				& C_{c}(i) = \frac{|V|-1}{\sum_{v \in V}d(i,v)}; &
				& C_{c}(j) = \frac{|V|-1}{\sum_{v \in V}d(j,v)} 
				\end{aligned}$\\
				\midrule
				Flow-rate\tnote{5}        & 
				$\begin{aligned}
				& Q_{i} = \sum_{j\in M(i)}\text{abs}(q_{ji})
				\end{aligned}$ & 
				$\begin{aligned}
				& Q_{ij} = \text{abs}(q_{ij})
				\end{aligned}$\\
				Weighted flow-rate &  
				$\begin{aligned}
				& \bar{Q}_{i} = \sum_{j\in M(i)}C_{b}(ij)\text{abs}(q_{ji})
				\end{aligned}$ & 
				$\begin{aligned}
				& \bar{Q}_{ij} = C_{b}(ij)\text{abs}(q_{ij})
				\end{aligned}$\\
				\bottomrule
			\end{tabular}
			\label{tab:topo_funcs}
			\begin{tablenotes}\footnotesize
				\item[1] $i,j,v \in V$ are nodes in network $G$. $(i,j)\in E$ are links. $a_{i,n}=1~\text{if}~(i,n)\in E~\text{else}~0$. 
				\item[2] $\sigma_{st}$ is the number of all shortest paths in $G$. $\sigma_{st}(i)$ and $\sigma_{st}(i,j)$ are number of shortest paths passing through $i$ and $(i,j)$, respectively.
				\item[3] $\lambda$ is a constant equal to the largest positive element in the eigenvector.
				\item[4] $d(i,v)$ is the shortest distance between $i$ and $v$ nodes if a path exists between them.
				\item[5] $q_{ji}$ is the maximum daily flow-rate from $j$ to $i$ during normal operation. $M(i)$ is the set of neighbor nodes of $i$.
			\end{tablenotes}
		\end{threeparttable}
	\end{sidewaystable}
	
	\subsection{Dimensionality reduction and development of prediction models}
	We implement dimensionality reduction using an iterative clustering algorithm introduced in this study. The iterative clustering algorithm combines clustering methods with regression algorithms to produce concise infrastructure resilience prediction models. This step consists of two subtasks: construction of cluster-level features and development of ML models.
	
	\subsubsection{Construction of cluster-level features from component-level features}
	Once the infrastructure components are categorized into different clusters, the next step is to derive the cluster-level features from the simulation dataset. In the dataset generated using \textit{InfraRisk}, the component-level information to be incorporated in the ML model is their initial functional states (disrupted or operational) after the occurrence of a disaster event.
	If a node cluster and a link cluster for infrastructure system $K$ generated by the clustering algorithm are denoted by $L_{K}^{v}:L_{K}^{v}\subseteq V_{K}$ and $L_{K}^{e}:L_{K}^{e} \subseteq E_{K}$, respectively, then the cluster-level features corresponding to the initial functional states of infrastructure components are derived as in Equation~\ref{eq:cluster_status}.
	
	\begin{equation}
	\delta_{i}^{L} = 
	\begin{cases} 
	1 & \text{if}~i\in L, \\
	0 & \text{otherwise.}
	\end{cases}\\
	\label{eq:cluster_status}
	\end{equation}
	where $\delta_{i}^{L}$ is the indicator of whether a component $i$ belongs to the cluster $L$ or not.
	
	Let $F^h$ denotes the set of cluster-level feature values corresponding to a topological or functional characteristic of the interdependent infrastructure network $\mathbb{K}$ and hazard $h$. The cluster-level feature $f_{L}^{h}:f_{L}^{h} \in F^{h}$ representing the initial disaster impact in the cluster $L$ is computed as the total number of failed components that belong to it (Equation~\ref{eq:cluster_feature}).
	
	\begin{equation}
	f_{L}^{h} = \sum_{\forall i\in \left\{V_{K}^{h},E_{K}^{h}: K\in \mathbb{K}\right\}}\delta_{i}^{L}.
	\label{eq:cluster_feature}
	\end{equation}  
	
	\subsubsection{Development of ML prediction models}
	The last step of our method is to develop the prediction models by employing state-of-the-art ML algorithms. Since the target variable (weighted EOH) in this study is continuous, we consider only regression algorithms. Commonly used regression algorithms include multiple linear regression (MLR), support vector regression (SVR), decision tree (DT), Random Forest (RF), and neural network regression (NN). For detailed discussions and applications of the major ML regression algorithms, we refer the readers to \cite{Berk2008, James2013}.
	
	Finally, we build the ML models using cluster-level features and recovery strategy as the predictors (Equation~\ref{eq:ml_model}).
	\begin{equation}
	\overline{\gamma} = \Phi\left ( \mathbb{F}, \mathbb{P} \right )
	\label{eq:ml_model}
	\end{equation}
	where $\mathbb{F} = \{F^{h}:h\in H\}$, $\mathbb{P} = \{\mathbb{P}_{h}:h\in H\}$, and $\Phi(\cdot)$ is the ML algorithm. To obtain robust and accurate models, cross-validation and hyper-parameter tuning are performed. 
	
	Two performance metrics are used to evaluate the goodness-of-fit of the ML models. The first metric is the coefficient of determination ($R^{2}$), which is defined as the proportion of the variation in the target variable captured by the prediction model (Equation~\ref{eq:R-squared}).
	\begin{equation}
	R^{2} =  1 - \frac{\sum_{i}(y_{i} - \hat{y}_{i})^{2}}{\sum_{i}(y_{i} - \bar{y})^{2}}
	\label{eq:R-squared}
	\end{equation}
	where $y_{i}$ is the $i$th observed value, $\hat{y_{i}}$ is the $i$th predicted value, and $\bar{y}$ is the mean  observed value.
	
	The second performance metric is the root mean square error ($RMSE$), which is the standard deviation of the prediction errors in the model (Equation~\ref{eq:RMSE}).
	\begin{equation}
	RMSE = \sqrt{\sum_{i}(y_{i} - \hat{y}_{i})^{2}}
	\label{eq:RMSE}
	\end{equation}
	The goal is to have a high $R^2$ and low $RMSE$ values. The unit of $RMSE$ in the this study is system performance-hours.
	
	A major aspect that is not yet resolved in the method is determining the optimal number of clusters in each infrastructure system for building the final ML model. Since an increase in the number of clusters would enhance the capability of the model to capture the spatial and network structure characteristics, an improvement in the quality of model prediction is expected. The optimal cluster count in each infrastructure system is determined using the {\em elbow} method \citep{Yuan2019} and our proposed iterative clustering method.
	
	The elbow method is an unsupervised method in which the sum of squared distances between observations and the centroids of the clusters they belong to is used as the performance measure for evaluating the consistency of clusters. The `elbow' of the curve connecting the sum of squared distances and total cluster count is determined, and the corresponding number of clusters in each infrastructure system is identified.
	
\begin{figure}[t!]
		\centering
		\includegraphics[width = .8\textwidth, trim = {0 0cm 0 0cm}, clip]{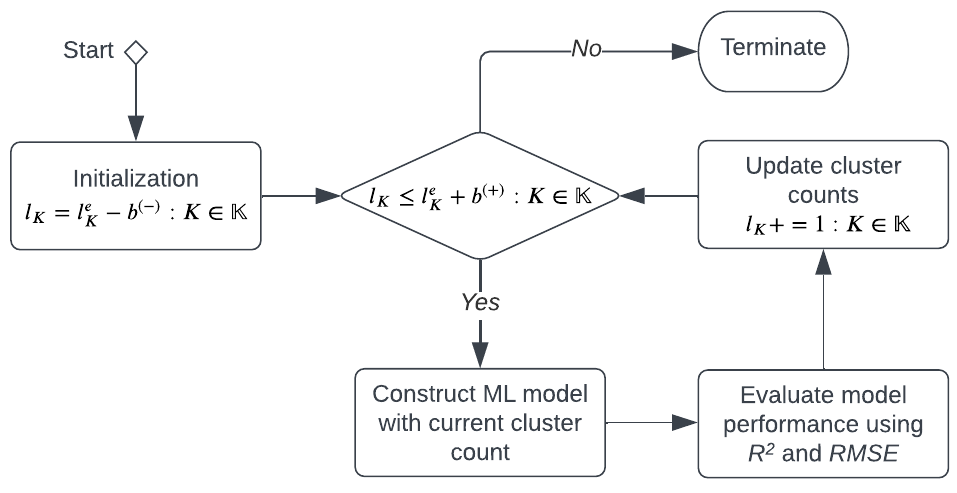}
		\caption{Iterative clustering method}
		\label{fig:iter_clust_algo}
	\end{figure}

	The second method proposed in this study to identify the optimal number of clusters in each infrastructure network is an iterative clustering algorithm (Figure~\ref{fig:iter_clust_algo}). It is a supervised method in which the performance of the ML model on test dataset is used as the performance measure. In this method, setting the ML model corresponding to the elbow method as the base model, we subsequently increased or decreased an equal number of clusters in all infrastructure systems to develop additional ML models. Once we had an adequate number of cluster combinations and corresponding ML models, we evaluated the improvement to the model goodness-of-fit ($R^{2}$) due to an increase in the number of clusters, and then adopt the most efficient model, i.e., ML model with highest $R^2$.
	
	Consider $\{l_{K}: K\in\mathbb{K}\}$ are the cluster counts corresponding to the infrastructure systems in the interdependent network. If the cluster counts determined by the elbow method are $\{l_{K}^{e}: K\in\mathbb{K}\}$, the maximum number of clusters that can be removed simultaneously from each network is $b^{(-)} =min\{l_{K}^{e}-1: K\in\mathbb{K}\}$, and the maximum number of clusters that can be added is $b^{(+)} = min\{|V_{K}| + |E_{K}| - l_{K}^{e}:K\in\mathbb{K}\}$. In the iterative clustering algorithm, we start by building the initial model with cluster counts $\{l_{K}^{e} - b^{(-)}: K\in\mathbb{K}\}$ and evaluating the model train and test performance metrics. In the subsequent iterations, we repeat the procedure after updating the cluster counts $\{l_{K}^{e} + b: K\in\mathbb{K}\}$, where $b \in \{-b^{(-)},\dots,b^{(+)}\}$. Finally, the optimal cluster counts in infrastructure networks are obtained by finding the `knee' of the curve between the test dataset $R^{2}$ and the total cluster count. We employ the {\em kneedle} algorithm \citep{Satopaa2011} for this purpose.
	
	\section{Case Study}
	\label{sec:case_study}
\begin{figure}[htbp]
		\centering
		\includegraphics[width = 0.85\textwidth, trim = {0 0 0cm 0cm}, clip]{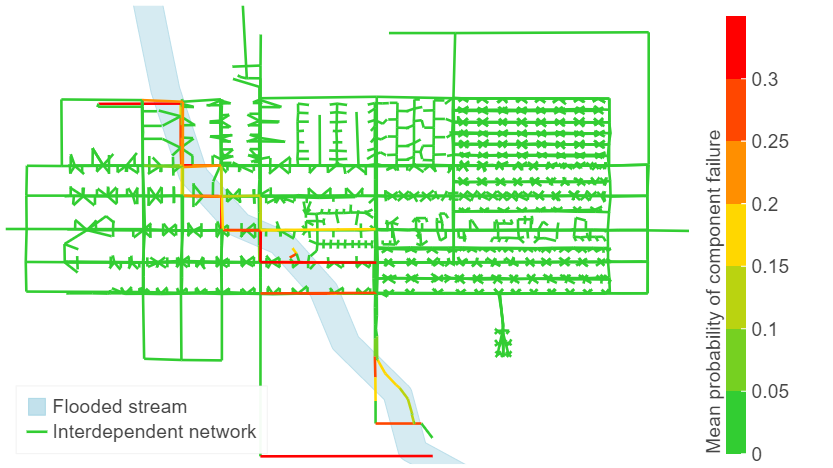}
		\caption{Micropolis network exposed to floods}
		\label{fig:micropolis_floods}
	\end{figure}

	The proposed methodology is implemented to develop a resilience prediction model for the Micropolis interdependent infrastructure network. Micropolis is a virtual city designed for 5000 inhabitants with water, power, and road networks \citep{Brumbelow2007}. We used hazard module in \textit{InfraRisk} to generate synthetic flood events and fail infrastructure components randomly based on the disaster intensity. A total of 325 flood scenarios are generated, assuming infrastructure components closer to the Micropolis stream are more likely to fail from a hazard (Figure~\ref{fig:micropolis_floods}). Each disaster scenario results in the failure of a specific set of infrastructure components in the interdependent infrastructure network. For this case study, we considered only water mains, power lines, and road links for failure as they are the most critical to the functioning of the respective infrastructure systems. It is found that 52 water mains, 22 power lines, and 17 road links along the water stream are either located or traversing through the regions exposed to the simulated floods. We limited the maximum number of failures in each flood scenario to 35 components to reduce the computational effort required for the case study. The disruptions to water links are modeled as leaks/pipe breaks, whereas that of power lines and road links are modeled by isolating them from the network.
	
	We assumed that each failed pipeline is remotely isolated by a predefined set of shutoff valves 10 minutes after the disaster occurred (considering sensing and actuation times). By isolating the leaking pipelines, the loss of water is minimized; however, isolating some segments of the water system would cutoff consumers located within the isolated regions. Once a water pipe is repaired, the corresponding isolating valves are opened, conditional upon whether that would interfere with the remaining repair actions. Similarly, when a power line is fully repaired, the corresponding circuit breakers are closed to allow electric power to flow through the line. In the case of damaged road links, each link is added back to the network after repair, and then the traffic assignment model recomputes the traffic flows.
	
	In this case study, each infrastructure system is assigned a repair crew for performing the post-disaster recovery. For implementing network recovery by component repair, three repair strategies are considered as follows:
	\begin{itemize}
		\item Betweenness centrality-based: Those components with a higher value of betweenness centrality are repaired first.
		\item Maximum flow-based: Those components that handle larger resource flow rates are repaired first.
		\item Zone-based: Components are repaired based on the zone in which they are located. The zones are prioritized in the order of central business district, industrial, and residential areas.
	\end{itemize}
	
	The recovery model in \textit{InfraRisk} also takes the accessibility to disrupted components into consideration and dynamically modifies repair sequences during the simulation.
	
\begin{figure}[b!]
		\centering
		\begin{subfigure}[h]{0.65\textwidth}
			\centering
			\fbox{\includegraphics[width = \textwidth]{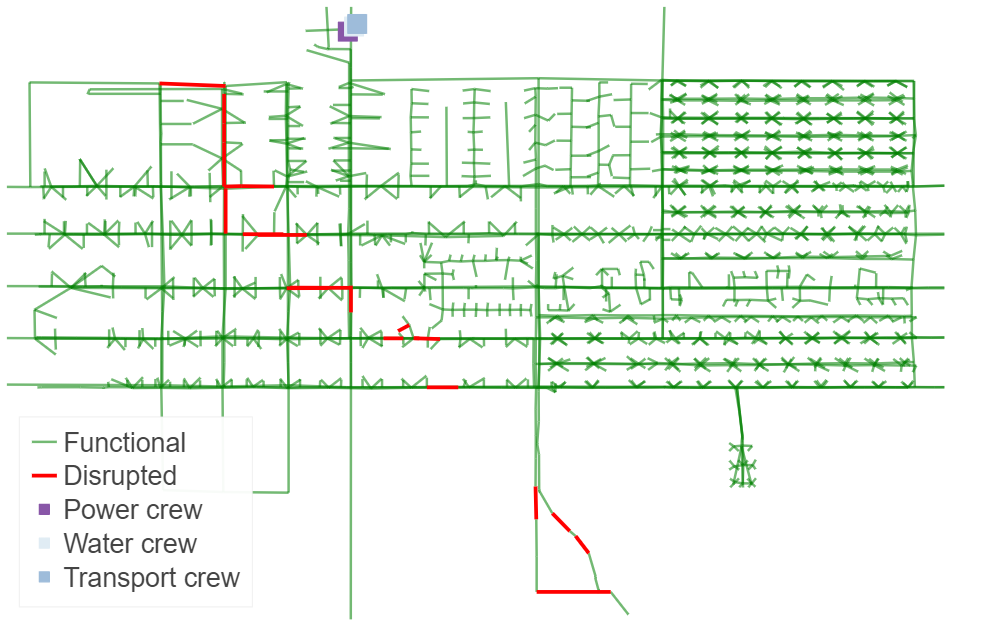}}
			\caption{Physically disrupted components and crew locations}
			\label{fig:disrupted_compons}
		\end{subfigure}
		
		\begin{subfigure}[h]{0.65\textwidth}
			\centering
			\includegraphics[width=\textwidth]{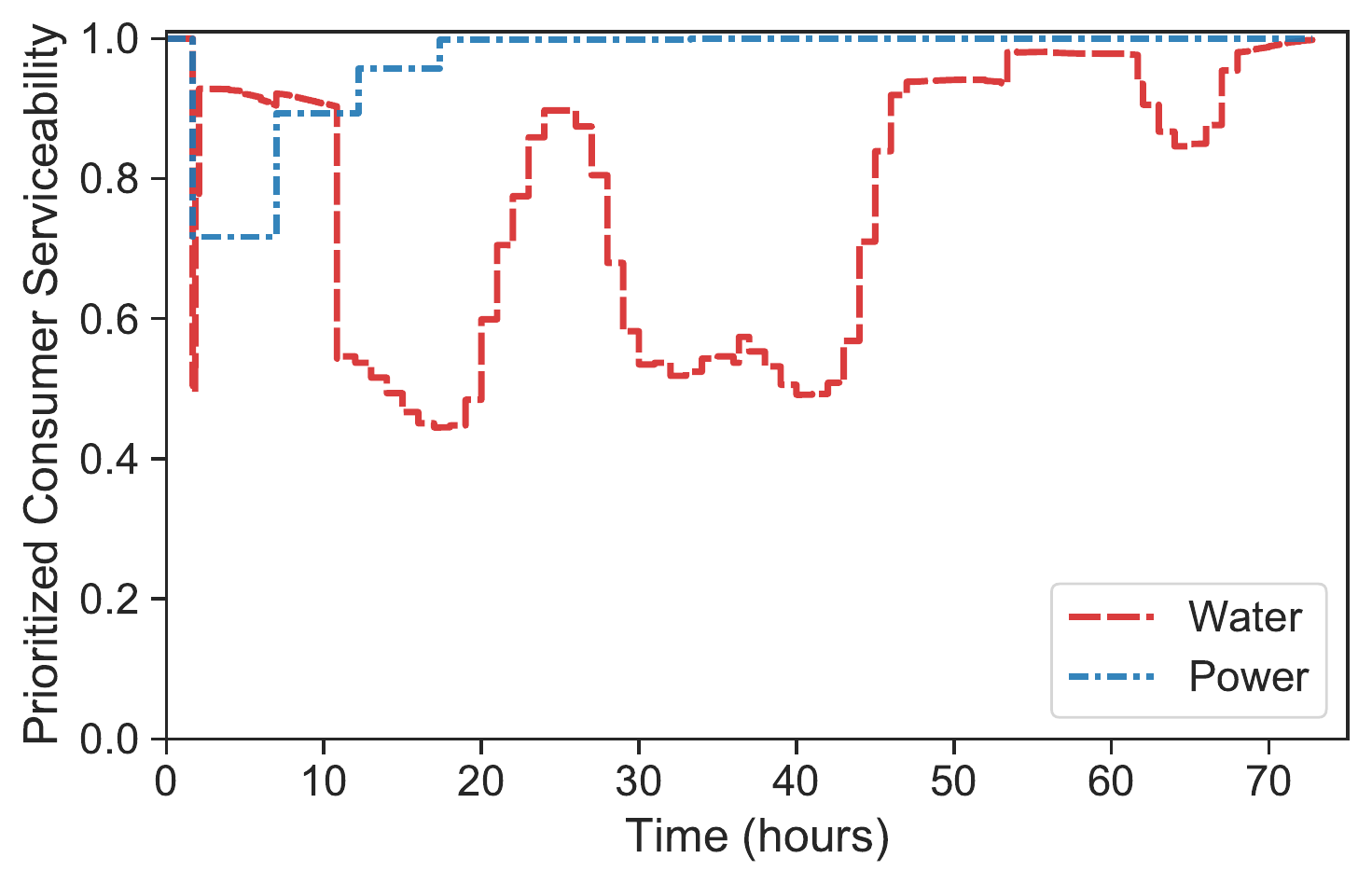}
			\caption{System recovery curves}
			\label{fig:pcs_plot}
		\end{subfigure}
		\caption{Simulation results from a simulated flood event}
		\label{fig:scenario_data}
	\end{figure}

	Figure~\ref{fig:scenario_data} presents the simulation results corresponding to one of the 325 simulated flood events. The flood event resulted in the failure of 14 water mains, six power lines, and four road links (Figure~\ref{fig:disrupted_compons}). 
	Figure~\ref{fig:pcs_plot} shows the performance of the Micropolis water and power network during the flood event when the capacity-based strategy is chosen for the network recovery. The infrastructure system performance is measured using the MOP in Equation~\ref{eq:pcs}. Both performance curves follow the typical resilience triangle used to characterize system resilience. The power system is restored to the pre-disaster state in approximately 33 hours, whereas the water crew takes approximately 68 hours to complete all the repair actions. The road network is fully restored in approximately 51 hours. Using Equation~\ref{eq:system_eoh}, the equivalent outage duration (in hours) corresponding to the disaster scenario in the water network is estimated to be 17.55 system performance-hours, and that in the power network is 2.92 system performance-hours.
	
\begin{figure}[b!]
		\centering
		\begin{subfigure}[h]{0.65\textwidth}
			\centering
			\fbox{\includegraphics[width=\textwidth]{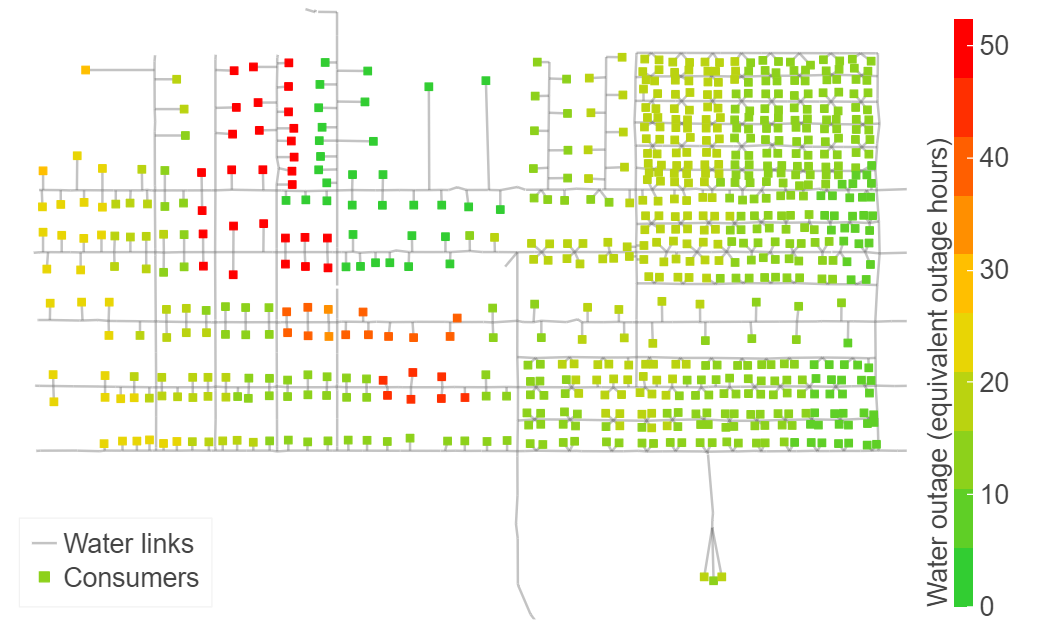}}
			\caption{Consumer-level equivalent water outage hours}
			\label{fig:water_eoh}
		\end{subfigure}
		\begin{subfigure}[h]{0.65\textwidth}
			\centering
			\fbox{\includegraphics[width = \textwidth]{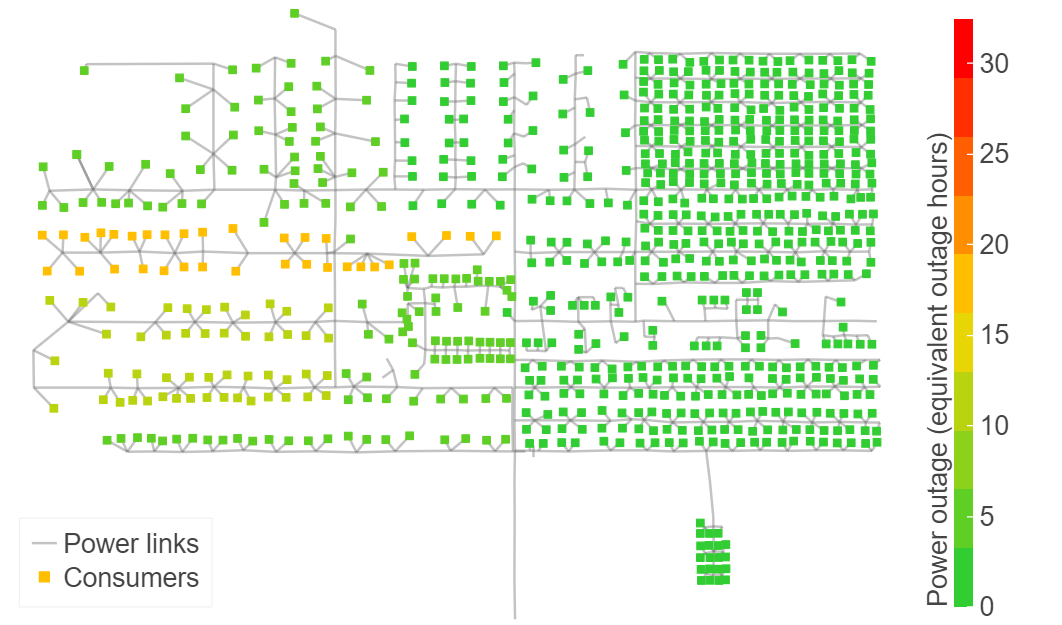}}
			\caption{Consumer-level equivalent power outage hours}
			\label{fig:power_eoh}
		\end{subfigure}
		\caption{Consumer-level impacts due to the simulated flood event}
		\label{fig:consumer_data}
	\end{figure}

	The consumer-level outages in power- and water utility services during the flood event are shown in Figure~\ref{fig:consumer_data}. 
	Even though the consumers in the adjacent areas along the flooded stream are affected by water outages the most, consumers in other parts of the network (especially those in the western and central Micropolis) are also affected by water outages (Figure~\ref{fig:water_eoh}). The leakage through failed pipes resulted in a reduced water head in the tank. The closure of shutoff valves isolated many consumers in other regions even though they were not directly affected by the flood event. On the other hand, the power outage is less severe compared to the water outage and is limited to the western side of the flooded stream (Figure~\ref{fig:power_eoh}). The drop in resilience values is mainly attributed to the consumers in downstream of failed power lines or opened circuit breakers who are disconnected from the rest of the power network.
	
\begin{figure}[htbp]
		\centering
		\includegraphics[width = 0.65\textwidth, trim = {0 0 0cm 0cm}, clip]{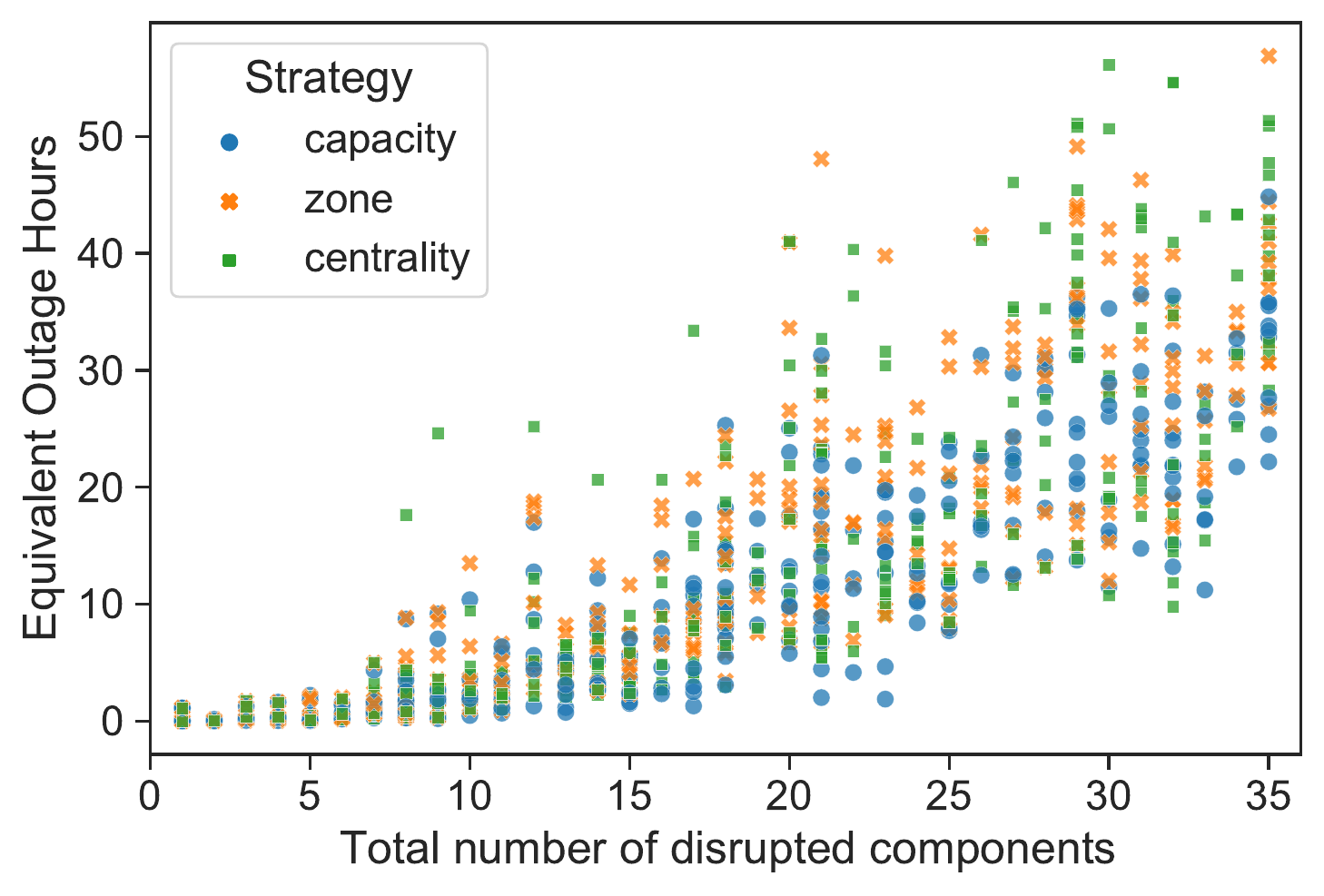}
		\caption{Weighted equivalent outage hours versus failure count}
		\label{fig:simulation_data}
	\end{figure}

	Next, the network resilience metrics (weighted equivalent outage hours) corresponding to all the disaster scenarios are calculated by assigning equal weights of 0.5 in Equation~\ref{eq:wEOH}. Only water and power systems are considered for evaluating the network resilience. Figure~\ref{fig:simulation_data} presents the distribution of the resilience metrics and their relationship with the recovery strategy adopted and the number of physically disrupted components. The results show a positive correlation between the weighted equivalent outage hours and the number of initially failed components.
	
	Next, ML models to predict the resilience (in terms of weighted equivalent outage hours) are developed as in Equation~\ref{eq:ml_model}. We used the Random Forest algorithm for this case study because of its simplicity and robustness. The ML models are built using 75\% of the data (training dataset), and the rest 25\% of the data (test dataset) is used for validation. The hyperparameters corresponding to the maximum depth and the total number of trees in the Random Forest algorithm are tuned for each model using 3-fold cross-validation.
	
\begin{sidewaysfigure}[htbp]
		\centering
		\includegraphics[width = 0.85\textwidth, trim = {0 .25cm 0 0.25cm}, clip]{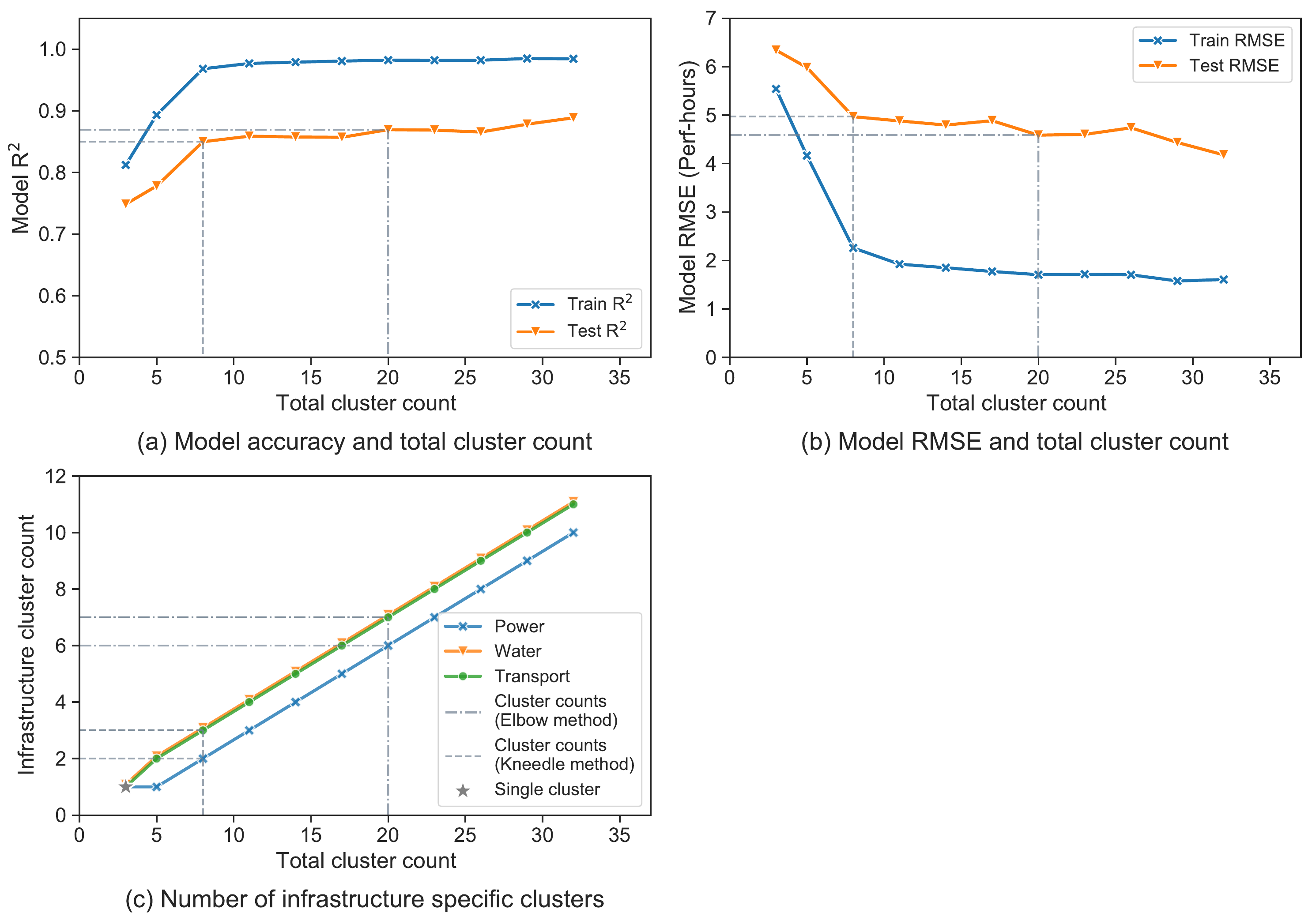}
		\caption{Relationship between model performance and number of infrastructure clusters.}
		\label{fig:elbow_knee}
	\end{sidewaysfigure}

	To identify the optimal clusters, we used the elbow method and the iterative clustering method combined with the {\em kneedle} algorithm. Figure~\ref{fig:elbow_knee} shows the results from the machine learning models developed using these methods. 
	
	In the case of the elbow method, the optimal number of clusters in water, power, and transport systems are found to be six, seven, and seven, respectively. The train- and test $R^{2}$ corresponding to the optimal cluster counts are 0.98 and 0.87, respectively. The corresponding $RMSE$ values are 1.71 and 4.59 system performance-hours.
	
	The machine learning models developed based on the iterative clustering algorithm reveal that the increase in the number of clusters in infrastructure systems initially leads to a noteworthy improvement in the model prediction. However, subsequent increases in the number of clusters only lead to marginal improvements in the same metrics. When the kneedle algorithm is used, the optimal number of clusters based on the iterative clustering method is found to be eight (two clusters in the power system and three each in the water system and the transport system). The corresponding Random Forest model has a train $R^{2}$ of 0.97 and a test $R^{2}$ of 0.85. The train $RMSE$ is 2.26 system performance-hours, and the train $RMSE$ is 4.97 system performance-hours.
	
	Both elbow and the iterative clustering method resulted in improved prediction models compared to that of the model with the single cluster-level feature for each infrastructure system. In the case of the elbow method, the relative improvement observed with the elbow method in test $R^{2}$ is 16.13\%, whereas that using the iterative clustering method is 13.52\%. At the same time, the models identified using the elbow method and the iterative clustering method resulted in significant reductions of 27.67\% and 21.65\% in test $RMSE$, respectively. The results show that the iterative clustering method resulted in a model with considerably fewer cluster features than the elbow method (eight cluster features compared to 20 cluster features) without compromising too much on the prediction accuracy. The summary of the models developed in this study is presented in Table~\ref{tab:model_details}.
	
	\begin{sidewaystable}[htbp]\small
		\centering
		\caption{Comparison of performance metrics of prediction models}
		\begin{tabular}{lccccccc}
			\toprule
			\multicolumn{1}{l}{\multirow{2}{2.25cm}{Clustering method}} & \multicolumn{1}{c}{\multirow{2}{1.25cm}{Total clusters}} & \multicolumn{1}{c}{\multirow{2}{1.5cm}{Train $R^{2}$}} & \multicolumn{1}{c}{\multirow{2}{1.5cm}{Test $R^{2}$}} &  \multicolumn{1}{c}{\multirow{2}{1.25cm}{Train $RMSE$}} & \multicolumn{1}{c}{\multirow{2}{1.5cm}{Test $RMSE$}} & \multicolumn{2}{c}{Relative change$^{\dagger}$}  \\
			\cmidrule{7-8}
			& & & & & & Test $R^{2}$ & Test $RMSE$ \\
			\midrule
			Single cluster   &       3  &        0.8121  &       0.7485  &   5.54 &  6.34    &  --  &        --\\
			Elbow method          &       20  &        0.9821  &       0.8693   &   1.71 &  4.59    &  +16.13\% &   -27.67\%\\
			Kneedle method            &       8 &        0.9680  &       0.8497  &   2.26 &  4.96    &  +13.52\% &   -21.65\% \\
			\bottomrule
		\end{tabular}
		\label{tab:model_details}
		\begin{tablenotes}\footnotesize
			\item[1] $^{\dagger}$All percentages are relative to the test dataset metrics obtained in the single cluster method.
		\end{tablenotes}
	\end{sidewaystable}
	
	\section{Conclusions}
	\label{sec:conclusions}
	In this study, we introduced infrastructure component clustering methods to generate concise infrastructure resilience prediction models with fewer features than the traditional models, thereby resolving the problem of high dimensionality. Disaster scenarios and resultant impacts on interdependent infrastructure networks are simulated using an interdependent infrastructure model. The disaster impacts on infrastructure are quantified using well-established resilience metrics. Prediction models are developed by applying machine learning algorithms. We clustered the component-level features into cluster-level features to reduce the number of features in the models (dimensionality reduction). The clusters are identified by partitioning infrastructure components with similar topological and functional characteristics as indicators of component vulnerability and importance. Finally, we proposed algorithms for determining the optimal number of infrastructure component clusters based on elbow- and iterative clustering methods.
	
	The clustering approach is a simple transformation technique that reduces the number of features (dimensionality reduction) and improves the model performance simultaneously. Since the clustering technique reduces the number of features in the model, improved prediction accuracy could be achieved with smaller simulation datasets. Therefore, the methodology can be adopted when simulation models' data generation is computationally expensive and time-consuming.
	
	The methodology could be further improved by considering the following aspects to produce more accurate prediction models.
	\begin{itemize}
		\item Along with topological and functional characteristics, simulation data may also be used to improve the quality of network partitioning.
		\item Clustering algorithms that implicitly learn the infrastructure network structure could produce more relevant clusters for resilience prediction.
		\item Interdependencies are currently not considered for clustering of infrastructure components.
		\item Additional component-level features relevant to resilience (for example, repair times) could be used in the clustering process to enhance the quality of clustering.
	\end{itemize}
	
	Though this paper only focused on interdependent infrastructure systems, our methodology can be used to design efficient ML models for any network problem where the characteristics of vertices or edges are treated as features. The proposed methodology could find applications to solve similar network problems in various fields, such as chemistry, medicine, finance, and social science.
	
	\section*{Acknowledgements}
	This research is supported by the National Research Foundation, Prime Minister’s Office, Singapore under its Campus for Research Excellence and Technological Enterprise (CREATE) programme.
	
	\section*{CRediT Author Statement}
	\noindent \textbf{Srijith Balakrishnan}: Conceptualization, Methodology, Data curation, Software, Visualization, Formal Analysis, Writing-Original Draft, Project administration, Supervision. \textbf{Beatrice Cassottana}: Conceptualization, Methodology, Validation, Resources, Writing-Review and Editing. \textbf{Arun Verma}: Conceptualization, Methodology, Validation, Writing-Review and Editing. 
	
	\section*{Data Availability}
	The \textit{InfraRisk} package can be downloaded from \href{https://github.com/srijithbalakrishnan/dreaminsg-integrated-model}{GitHub}\footnote{\href{https://github.com/srijithbalakrishnan/dreaminsg-integrated-model}{https://github.com/srijithbalakrishnan/dreaminsg-integrated-model}}. Documentation and codes for sample simulations are available in the \textit{InfraRisk} package. The simulation data and codes pertaining to this study can be shared upon request.
	
	\setlength{\bibsep}{0pt plus 0.3ex}
	{\small\bibliography{sample}}

\end{document}